\def\year{2019}\relax
\documentclass[letterpaper]{article} %DO NOT CHANGE THIS
\usepackage{aaai19}  %Required
\usepackage{times}  %Required
\usepackage{helvet}  %Required
\usepackage{courier}  %Required
\usepackage{url}  %Required
\usepackage{graphicx}  %Required

\usepackage{microtype}
\usepackage{subfigure}
\usepackage{amsmath}
\usepackage{amssymb}
\usepackage{amsfonts} 
\usepackage{diagbox}
\usepackage{booktabs}

\graphicspath{{figure/}}

\begin{document}

\title{Cognitive Deficit of Deep Learning in Numerosity}

\author{Xiaolin Wu $^{\dagger \star}$, Xi Zhang $^{\star}$, Xiao Shu $^{\dagger \star}$ \\
$^{\dagger}$ Department of Electrical \& Computer Engineering, McMaster University, Canada \\
$^{\star}$ Department of Electronic Engineering, Shanghai Jiao Tong University, China \\
xwu@ece.mcmaster.ca, zhangxi\_19930818@sjtu.edu.cn, shux@mcmaster.ca  
}

\maketitle

\begin{abstract}
  Subitizing, or the sense of small natural numbers, is an innate cognitive function of humans and primates; it responds to visual stimuli prior to the development of any symbolic skills, language or arithmetic. Given successes of deep learning (DL) in tasks of visual intelligence and given the primitivity of number sense, a tantalizing question is whether DL can comprehend numbers and perform subitizing.  But somewhat disappointingly, extensive experiments of the type of cognitive psychology demonstrate that the examples-driven black box DL cannot see through superficial variations in visual representations and distill the abstract notion of natural number, a task that children perform with high accuracy and confidence.  The failure is apparently due to the learning method not the CNN computational machinery itself. A recurrent neural network capable of subitizing does exist, which we construct by encoding a mechanism of mathematical morphology into the CNN convolutional kernels. Also, we investigate, using subitizing as a test bed, the ways to aid the black box DL by cognitive priors derived from human insight.  Our findings are mixed and interesting, pointing to both cognitive deficit of pure DL, and some measured successes of boosting DL by predetermined cognitive implements.  This case study of DL in cognitive computing is meaningful for visual numerosity represents a minimum level of human intelligence.
\end{abstract}

\section{1. Introduction}

\subsection{1.1 Background and motivation}

In the past decade deep neural networks have rapidly developed into a powerful problem-solving paradigm that has found a wide gamut of applications, spanning almost all academic disciplines.  In particular, deep convolutional neural networks (DCNNs) are lauded for their apparent visual intelligence, by which we refer to the successes enjoyed by DCNNs in visual pattern analysis, recognition and classification tasks \cite{AlexNet,VGG,GoogleNet,ResNet,Tang2014,FaceNet,Deepid3}.   Although convolutional neural networks were originally inspired by the knowledge of animal cortex \cite{Hubel1968}, our understanding of the inner working mechanism of DCNNs is outstepped or arguably even shrouded by their functional prowess in many applications.

In this work we are intrigued by and try to understand the uncomforting contrast between the abundance that DL can accomplish in visual recognition tasks, some of which are quite challenging even for humans (e.g., judging if two face images are of the same person \cite{GaussianFace}), and how severely handicapped it becomes when tested on cognitive tasks as simple and basic as learning the concept of numbers.
We conduct a family of cognitive experiments to test if DL can, under various levels of supervision, learn the simple concept of natural numbers by observing sample images containing a varying number of objects in different positions, orientations, sizes, shapes and colors.

The awareness of numbers, or numerosity perception is a neurocognitive function possessed by human infants prior to speech and any symbolic learning and even by animals \cite{Reas,Harvey,Brannon,Burr,Dehaene,Nieder,Xu}.  Furthermore, numerosity perception is innate very much like taste, sight, touch, smell and sound, although it seems to be a higher order cognitive construct than the common five senses.  Arguably, numerosity is the simplest task of cognitive computing and thus it serves as an ideal Turing-type test to challenge whether DL, or any AI machinery for that matter, can match humans.

\subsection{1.2 Related work}

Stoianov and Zorzi proposed a forward neural network of two hidden layers to learn numerosity estimation \cite{nature2011}.  The learnt neural network model was found to exhibit a numerosity estimation behavior.  However, the authors did not push far enough when testing their neural network for cognitive power in numerosity.  The synthetic objects of test images do not vary in shape, color, or orientation; the size variation of objects remain in the same range of the training images.  As such, the success of the neural network in making statistical inference under the i.i.d.~condition is expected.  In addition, the numerosity discussed in \cite{nature2011} did not include subitizing (the precise judgement of small numbers) that is a key topic in our study.

Very recently, Ritter et al.~investigated the interpretability problem in DCNNs using the methods of cognitive psychology \cite{Ritter}. They found that one shot learning methods trained on ImageNet have a human-like bias when associating a class of objects with a word or label.  The authors advocate to leverage tools of cognitive psychology to better understand DCNNs.  Many of our experiments to test cognitive power of DL are also  partly inspired by works of cognitive psychology.

\subsection{1.3 Paper outline and contributions}

The main enquiry of this work is whether the black box DL can reach the level of abstraction and reasoning of humans, beyond i.i.d.~statistical inference.  After all, not all mental processes fall into the realm of statistical inferences; human brain is more than just a Bayesian machine.  A key finding of this paper is that the end-to-end black box DL fails the numerosity tests, although the number sense represents a minimum level of intelligence, far more primitive than natural language understanding.
Hence, this study adds a fresh anecdote to the widely held critiques on the deficit of DCNNs in cognitive computing.  The failures of DCNNs in our cognitive experiments are analyzed.  The analysis exposes the overreliance of existing DL methods on sample statistics at the expense of scene semantics. DL is easily confused by immaterial variations of the same visual representation of numbers, and fails to generalize in the size, shape and color of objects.

In sharp contrast to the failures of the black box learning, by incorporating a mechanism of mathematical morphology into convolutional kernels, we are able to construct a recurrent convolutional neural network (RCNN) to subitize deterministically rather than statistically. The proposed RCNN is neurocognitively motivated and has a simple and small core of very few parameters.
This compact model can abstract the number concept from wide visual signal variations, which the pure data-driven DL fails to accomplish.  The only caveat is that the proposed RCNN has the benefit of human-knowledge.

Another contribution of this paper is to make deep learning break the i.i.d.\ limitation by head starting the black box learner on cognitive primitives derived from human insight.  We dissect the proposed RCNN for subitizing into subnetworks for morphological erosion, connected components and counting.  These subnetworks can be separately trained and combined to learn the subitizing task with a generalization ability beyond the i.i.d.\ limitation.  If considering the above task-specific subnetworks as preexisted (innate) cognitive implements prior to machine learning, then the above outlined enhanced learning approach parallels the learning model of "nurturing the nature" in cognitive psychology.

The remainder of this paper is structured as follows.
Section 2 is a brief review of the existing knowledge on human visual numerosity perception.  Section 3 presents the designs, results and discussions of our experiments on the capability of DCNNs to learn subitizing.  After exposing the difficulties of DCNNs in abstracting numbers from training images, we introduce in Section 4 a boundary representation to improve the abstraction power.  In Section 5, we design and discuss the deterministic RCNN that can perform subitizing even when images fall outside the distribution of the training data, and this success is contrasted with the black box DL approach.  In Section 6, we investigate how to boost DL and improve its generalization capability by using preexisted cognitive implements. Discussion and conclusions are in Section 7.

\section{2. Human Visual Numerosity}

Neuroscientific evidence and data exist to suggest that children are endowed with, prior to speech and symbolic learning, the aptitude of number appreciation\cite{Dehaene1999}.  Neural circuits dedicated to numerical cognition are found \cite{Harvey}.
The evolutionary values of being keenly aware of number are self-evident, including the advantages in foraging \cite{Krebs1987}, reproduction \cite{Lyon2003}, and social activities \cite{Mccomb1994,Wilson2002}.

Visual numerosity involves two neurocognitive processes: subitizing and approximate numerosity.  Subitizing refers to the human ability, discovered by Kaufman \cite{Kaufman}, for rapid, accurate, and confident judgment of the number of items in a small set. Approximate numerosity refers to the instinct of human and other species to spontaneously estimate the number of items in an environment \cite{Brannon2003}.

Opposite edges of an area in the right superior parietal lobe are identified to maximally respond to small and large quantities \cite{Harvey}.  The estimation precision of approximate numerosity mechanism obeys Weber's law like hearing, seeing, tasting and other basic sensory functions.  The populations of two sets of objects can be reliably ranked only if the two numbers differ by a sufficiently large ratio governed by Weber's law \cite{Fechner2012}.

Subitizing is another cognitive function; it infers the exact number of objects by tracking individual objects in space and time.  The cognition relies on the spatio-temporal coherence: parts of an object stay as a bounded whole in both space and time, reminiscing on the notion of the connected compound in mathematics and digital images.  Studies show that the togetherness of an object is how human infants perceive object boundaries \cite{Spelke}.  However, the reliability of subitizing is limited to small numbers.  If the number exceeds four, then the approximate numerosity mechanism takes over the task of numerical cognition \cite{Nieder2004,Piazza2004,Tokita2010,Whalen1999}.

\section{3. Visual Numerosity of Deep Convolutional Neural Networks}

DCNNs can be trained to count a specific type of objects in a particular environment, such as the pedestrians on a street or cells under a microscope \cite{Zhang2015,Xie2016}.  But these DL methods cannot generalize, like humans, across different objects in different backgrounds.  Numerosity requires the abstraction of natural numbers that are disassociated from, say, 2 faces or 3 cells.  The ability, or lack of it, of deep learning to understand numbers is a probing case study to compare humans and AI machines in cognitive power, 
as numerosity is arguably one of the most basic forms of human intelligence.

Numerosity relies on the topological feature of connected component, rather than geometrical features such as size, shape, spatial arrangement of the objects, etc.  It will be tantalizing to find out whether DCNNs can be taught to ``see through'' superficial geometric variations and understand the abstract notion of numerosity.  In this research our goal is not to solve practical counting problems, but rather examine the cognitive potential of data-driven black box DCNNs.

\begin{figure}[!ht]
    \centering
      \includegraphics[width=1.2cm]{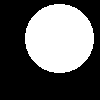}
      \includegraphics[width=1.2cm]{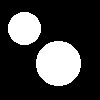}
      \includegraphics[width=1.2cm]{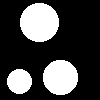}
      \includegraphics[width=1.2cm]{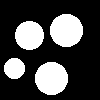}
      \includegraphics[width=1.2cm]{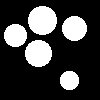}
      \includegraphics[width=1.2cm]{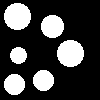} \\
      $n$=1 \hskip 0.6cm $n$=2 \hskip 0.6cm $n$=3 \hskip 0.6cm
      $n$=4 \hskip 0.6cm $n$=5 \hskip 0.6cm $n$=6 \hskip 0.6cm\\
      \caption{Sample training images for class $n$ ($1 \leq n \leq 6$).}
    \label{fig:s_t_circle}
\end{figure}

In cognitive science numerosity is a raw perception rather than resulting from arithmetic, accordingly we model it as a classification system. In case of subitizing, the system reads visual representations of small numbers and emits class labels, each class for a different number.  We train a DCNN for subitizing by teaching it to perform the following 6-label classification task.  The 6 output labels correspond to natural numbers 1 through 6.
The training images for class $n$ ($n=1,2,\cdots,6$) consist of $n$ white solid circles in black background (see Fig.~\ref{fig:s_t_circle}).  These circles are of random sizes, and furthermore the number of circles in a training image is statistically independent of the total area of these circles.

To focus on the cognitive aspect of our experiments, the visual representations of numbers for training the DCNN classifier are made as simple as possible: synthetic, noise-free simple objects are randomly placed against a constant featureless background.  Striping from the training images the intricacies of real-world application problems is to guide the DCNN classifier towards the discovery of the invariant of all sample images in each class, that is, the number of connected components.  Here the connected component can be considered as the minimal and hence most robust feature for the cognitive task of subitizing.  Moreover, the connected component has a biology basis for number sense because it encodes the spatial togetherness of an object \cite{Spelke}.  As DCNNs have a tendency to draw statistical inferences, which is a detour to our cognitive task, we make the number of circles in a training image statistically independent of the total area of these circles.

The subitizing DCNN consists of five convolutional layers and two fully-connected layers.
The detailed network configuration is outlined in Table~\ref{table:dcnn}.

\begin{table}[!ht]
  \centering
  \caption{The subitizing DCNN configuration. The convolutional layer parameters are denoted as ``Conv-$\langle$filter size$\rangle$-$\langle$number of filters$\rangle$''.}
  \label{table:dcnn}
  \begin{tabular}{lcc}
    \toprule
    Layers & Maxpooling \\
    \midrule
    Conv-7-32 & Yes \\
    \midrule
    Conv-5-32 & Yes \\
    \midrule
    Conv-3-64 & No \\
    \midrule
    Conv-3-64 & No \\
    \midrule
    Conv-3-128 & Yes \\
    \midrule
    FC-128 & - \\
    \midrule
    FC-6 & - \\
    \midrule
    Softmax & - \\
    \bottomrule
  \end{tabular}
\end{table}

Given the past successes of DCNNs in solving visual classification problems, it is not surprising that the above DCNN can perform subitizing almost perfectly on test images of circles that are drawn from the same distribution of the training images.  But the DL approach is incapable of a level of abstraction that is trivial for children on the task of subitizing.  The experimental results in the following four experiments show that the DCNN gets easily confused by superficial variations in visual number representations, such as changes in the size, shape and color of the objects, and in region-boundary duality.  The accuracy of number judgement decreases greatly when the test images deviate from the training images only in form not in essence.  Again, this failure is expected because the consensus is that DL can only work under the i.i.d.\ condition.

% \vspace{0.5ex}
\textbf{Experiment 1.  Abstraction from object sizes}

The test images are the same as the training images, only the circles are of $50\%$ greater variation in size than the circles in the training image set.

\textbf{Results}: As shown in Table \ref{table:set1}, a modestly increased
variability in object size significantly decreases the classification
accuracy of the subitizing DCNN.  Interestingly, very much like subisizing by humans, the
error of the trained subitizing DCNN remains very small up to 4, then jumps
for larger numbers.

\begin{table}[!t]
    \centering
    \caption{Performance of the subitizing DCNN on images
      of circles of $50\%$ greater size variation than the training
      set (the probability of perceiving $n$ as $m$).}
    \label{table:set1}
    \begin{footnotesize}
      \begin{tabular}{c|cccccc}
        \toprule
        \diagbox[width=1cm]{$n$}{$m$} & 1 & 2 & 3 & 4 & 5 & 6 \\
        \midrule
        1 & 1 & 0 & 0 & 0 & 0 & 0 \\
        2 & 0.003 & 0.997 & 0 & 0 & 0 & 0 \\
        3 & 0 & 0.010 & 0.990 & 0 & 0 & 0 \\
        4 & 0 & 0 & 0.041 & 0.959 & 0 & 0 \\
        5 & 0 & 0 & 0 & 0.328 & 0.672 & 0 \\
        6 & 0 & 0 & 0 & 0.001 & 0.45 & 0.549 \\
        \bottomrule
      \end{tabular}
     \end{footnotesize}
\end{table}

% \vspace{0.5ex}
\textbf{Experiment 2.  Abstraction from object shapes}

The subitizing DCNN is generalized to different shapes by replacing white
circles of the training images by white equilateral triangles, squares
and pentagons in test images.  

\textbf{Results}: Tables~\ref{table:set2.1} and \ref{table:set2.2}
clearly reveal a lack of generality of the trained subitizing DCNN to
object shapes.
In the case of equilateral triangles, the trained subitizing DCNN cannot even
correctly judge very small quantities, 1 to 3, which are well within
the threshold of subitizing for humans; also, it systematically
overestimates, with a low overall accuracy of 0.45.  For the test
images of squares, the accuracy of the subitizing DCNN increases to 0.76 on
average; it is above 0.8 only for numbers 1, 2 and 6, but quite low for 3, 4, and 5.

\begin{table}[!ht]
    \centering
    \caption{Performance of the subitizing DCNN on
      images of triangles (the probability of perceiving $n$ as $m$).}
    \label{table:set2.1}
    \begin{footnotesize}
      \begin{tabular}{c|cccccc}
        \toprule
        \diagbox[width=1cm]{$n$}{$m$} & 1 & 2 & 3 & 4 & 5 & 6 \\
        \midrule
        1 & 0.327 & 0.673 & 0 & 0 & 0 & 0 \\
        2 & 0.031 & 0.441 & 0.528 & 0 & 0 & 0 \\
        3 & 0.001 & 0.027 & 0.361 & 0.611 & 0 & 0 \\
        4 & 0 & 0.001 & 0.022 & 0.287 & 0.625 & 0.038 \\
        5 & 0 & 0 & 0 & 0.044 & 0.364 & 0.592 \\
        6 & 0 & 0 & 0 & 0.003 & 0.067 & 0.930 \\
        \bottomrule
      \end{tabular}
    \end{footnotesize}
\end{table}

\begin{table}[!ht]	
    \centering
    \caption{Performance of the subitizing DCNN on
      images of squares (the probability of perceiving $n$ as $m$).}
    \label{table:set2.2}
    \begin{footnotesize}
      \begin{tabular}{c|cccccc}
        \toprule
        \diagbox[width=1cm]{$n$}{$m$} & 1 & 2 & 3 & 4 & 5 & 6 \\
        \midrule
        1 & 0.876 & 0.124 & 0 & 0 & 0 & 0 \\
        2 & 0.019 & 0.811 & 0.170 & 0 & 0 & 0 \\
        3 & 0 & 0.009 & 0.641 & 0.350 & 0 & 0 \\
        4 & 0 & 0 & 0.005 & 0.686 & 0.309 & 0 \\
        5 & 0 & 0 & 0 & 0.020 & 0.549 & 0.431 \\
        6 & 0 & 0 & 0 & 0 & 0.022 & 0.978 \\
        \bottomrule
      \end{tabular}
    \end{footnotesize}
\end{table}

% \vspace{0.5ex}
\textbf{Experiment 3.  Abstraction from object colors}

The objects in this test set are statistically identical to the
objects in the training set in terms of geometry, but their foreground / background colors are swapped.

\textbf{Results}: As shown in Table~\ref{table:set3}, the trained subitizing DCNN completely fails the subitizing
test, although the test images only undergo a superficial systematic
change from the training images.  The average accuracy is only 0.42.

% \vspace{0.5ex}
\textbf{Experiment 4.  Abstraction from region-boundary duality}

In human vision, the region and boundary representations of objects
are patently dual of each other.  In Marr's theory of computational
vision \cite{Marr82}, the primal sketch of objects is a vital to cognition. 
We test the trained subitizing DCNN for subitizing with the boundary version of training
images, and examine how well it can generalize to the simple region-boundary duality.  
This test set consist of white rings of random size, position and orientation under black background.  
Clearly in this case, the test images, although carrying the same meaning in the notion of
numerosity, have drastically different statistics from the training set.  
As the trained subitizing DCNN tries to make a statistical inference, it
fails catastrophically on the cognitive task of subitizing (see Table~\ref{table:set4}).

\begin{table}[!t]
 	\centering
  	\caption{Performance of the subitizing DCNN on images of
    	swapped colors (the probability of
    	perceiving $n$ as $m$).}
  	\label{table:set3}
  	\begin{footnotesize}
	  \begin{tabular}{c|cccccc}
	    \toprule
	    \diagbox[width=1cm]{$n$}{$m$} & 1 & 2 & 3 & 4 & 5 & 6 \\
	    \midrule
	    1 & 0.160 & 0.800 & 0.040 & 0 & 0 & 0 \\
	    2 & 0.010 & 0.340 & 0.650 & 0 & 0 & 0 \\
	    3 & 0 & 0.040 & 0.680 & 0.280 & 0 & 0 \\
	    4 & 0 & 0.010 & 0.160 & 0.670 & 0.160 & 0 \\
	    5 & 0 & 0 & 0.060 & 0.500 & 0.420 & 0.020 \\
	    6 & 0 & 0 & 0 & 0.180 & 0.570 & 0.250 \\
	    \bottomrule
	  \end{tabular}
	\end{footnotesize}
\end{table}

\begin{table}[!ht]
  \centering
  \caption{Performance of the subitizing subitizing DCNN on images of
    white rings (the probability of perceiving $n$ as $m$).}
  \label{table:set4}
  \begin{footnotesize}
	  \begin{tabular}{c|cccccc}
	    \toprule
	    \diagbox[width=1cm]{$n$}{$m$} & 1 & 2 & 3 & 4 & 5 & 6 \\
	    \midrule
	    1 & 0.004 & 0.647 & 0.349 & 0 & 0 & 0 \\
	    2 & 0 & 0.002 & 0.420 & 0.578 & 0 & 0 \\
	    3 & 0 & 0 & 0.010 & 0.458 & 0.523 & 0.009 \\
	    4 & 0 & 0 & 0 & 0.096 & 0.576 & 0.328 \\
	    5 & 0 & 0 & 0 & 0.002 & 0.194 & 0.804 \\
	    6 & 0 & 0 & 0 & 0 & 0.011 & 0.989 \\
	    \bottomrule
	  \end{tabular}
   \end{footnotesize}
\end{table}

If the black-box DL approach is ever able to attain the level of human intelligence, then it has to generalize or reason beyond statistical inference under the i.i.d.\ condition.  In this context, we are interested in the possibility, or lack of it, to augment DL in any ways so that it can eventually distill the concept of natural numbers, a very basic cognitive construct for humans and primates, from a non-exhaustive set of visual examples.  Our investigations and developments along the above line of enquiry are presented in the following two sections.

\section{4. Generalization with Boundary Representation}

\begin{figure}[!t]
    \centering
    \includegraphics[width=1cm]{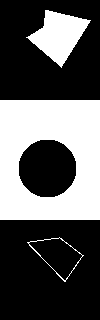}
    \includegraphics[width=1cm]{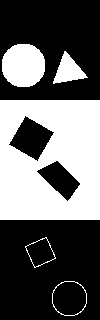}
    \includegraphics[width=1cm]{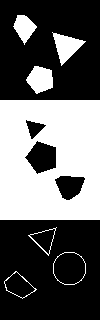}
    \includegraphics[width=1cm]{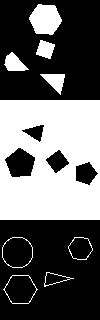}
    \includegraphics[width=1cm]{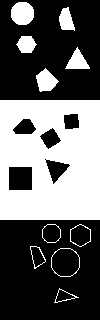}
    \includegraphics[width=1cm]{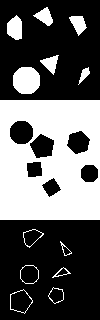}
    \caption{The sample images that contain objects of varied shapes,
      sizes, colors, and in either region or boundary
      representations.}
    \label{fig:s_random}
\end{figure}

Experiments 1 through 4 expose the inability of the DCNN to acquire the cognitive function of subitizing by generalizing to superficial variations in the size, shape, color and the region-boundary duality of the objects.  One way to make the DCNN generalize better is to enrich the training images and include objects of varied shapes, sizes, colors, and in either region or boundary representations, as shown in Fig.~\ref{fig:s_random}.  The objects are circles and simple $n$-gons, not necessarily convex, $1 \leq n \leq 6$, and in random configuration.

\begin{figure}[!t]
    \centering
    \includegraphics[width=1cm]{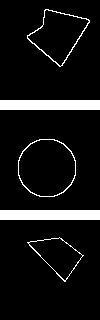}
    \includegraphics[width=1cm]{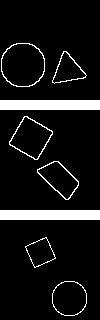}
    \includegraphics[width=1cm]{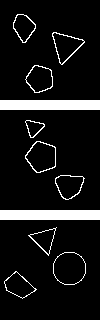}
    \includegraphics[width=1cm]{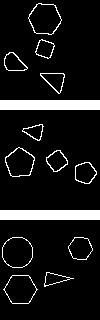}
    \includegraphics[width=1cm]{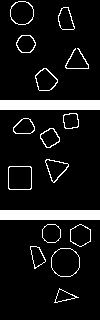}
    \includegraphics[width=1cm]{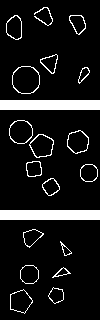}
    \caption{Unified boundary representations of the training images in
      Fig.~\ref{fig:s_random}.}
    \label{fig:s_random_process}
\end{figure}

More importantly, we boost the abstraction capability of the subitizing DCNN by allowing it to learn from the boundary representation of objects instead from the set of pixels.  The boundary representation, as illustrated in Fig.~\ref{fig:s_random_process}, is given to the DCNN as a preexisted primitive to associate numbers with objects.  Very much in analogy to the nature-nurture characterization in cognitive psychology, the boundary representation here is considered innate to the DCNN and it does not need to be learnt from raw data. As illustrated by comparing Fig.~\ref{fig:s_random} and Fig.~\ref{fig:s_random_process}, the boundary primitive unifies the presentations of varied example images, regardless whether the original training image has white objects in black background or reversed, or whether the objects are represented in solid color or boundary sketch.  This relieves the DCNN the burden of abstracting from different colors and region-boundary duality.

  % \hfill

\begin{figure}[!ht]
    \centering
    \includegraphics[width=7cm, height=5.5cm]{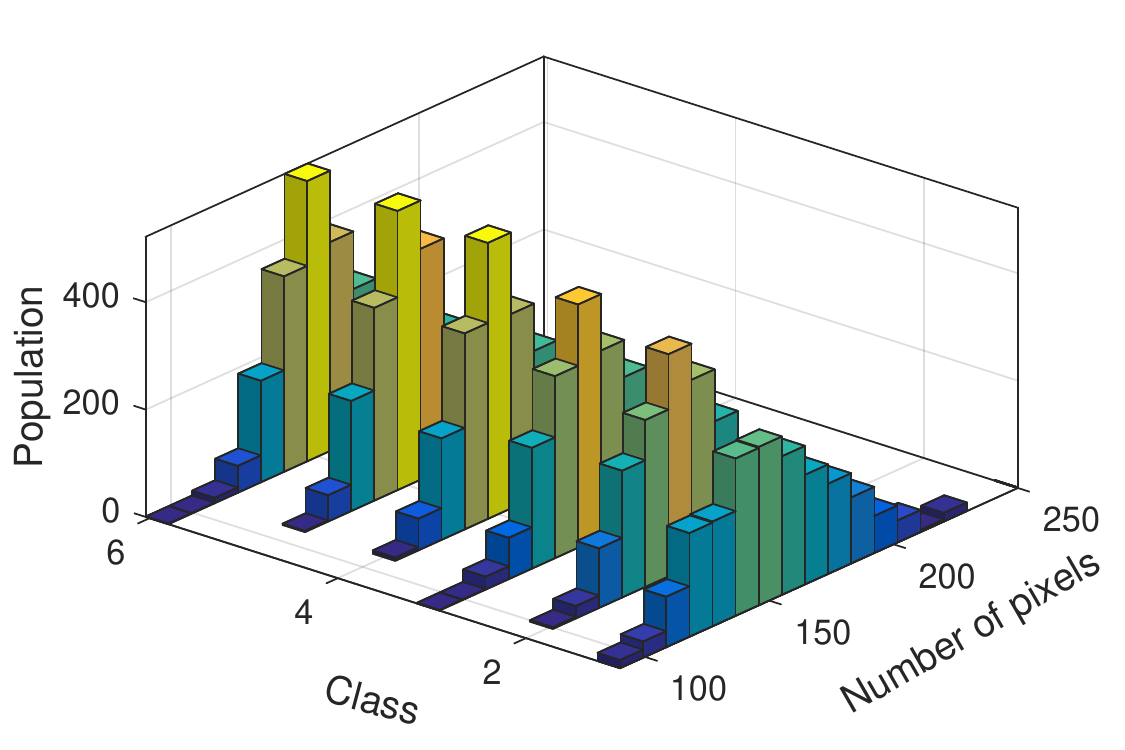}
    \caption{The histograms of the number of edge pixels $c$ for
      different classes $n$.}
    \label{fig:hist_area}
\end{figure}

Also, to prevent the DCNN from estimating the number of objects $n$ from the number of boundary pixels $c$ (the sum of the perimeters of all objects) in Fig.~\ref{fig:s_random_process}, and force it to discover via supervised learning the topological construct of connected component, which is the most essential and robust feature for numerosity, we normalize $c$ by scaling the objects such that $c$ has very close distributions for different classes ($n$), as shown in Fig.~\ref{fig:hist_area}.

Next, we retrain the 6-label DCNN classifier using the normalized boundary images, and examine if, after the extra helps are given towards generalization, the DCNN can subitize on an arbitrary visual representation of small natural numbers.  The experimental results of the human-guided DCNN on test images are tabulated in Table~\ref{table:random}.  The table shows very high classification rates of the newly trained DCNN on the normalized boundary images.  Indeed, if the object boundaries are extracted and provided to the DCCN, it appears to comprehend small natural numbers via an abstraction of different visual representations.

\begin{table}[!ht]
    \centering
    \caption{Performance of the human-guided DCNN on edge maps. (the probability of perceiving $n$ as $m$).}
    \label{table:random}
    \begin{footnotesize}
      \begin{tabular}{c|cccccc}
        \toprule
        \diagbox[width=1cm]{$n$}{$m$} & 1 & 2 & 3 & 4 & 5 & 6 \\
        \midrule
        1 & 1 & 0 & 0 & 0 & 0 & 0 \\
        2 & 0 & 1 & 0 & 0 & 0 & 0 \\
        3 & 0 & 0.020 & 0.970 & 0.010 & 0 & 0 \\
        4 & 0 & 0 & 0.020 & 0.930 & 0.050 & 0 \\
        5 & 0 & 0 & 0 & 0.040 & 0.790 & 0.170 \\
        6 & 0 & 0 & 0 & 0 & 0.080 & 0.920 \\
        \bottomrule
      \end{tabular}
    \end{footnotesize}
\end{table}

\begin{table}[!ht]
    \centering
    \caption{Performance of the subitizing DCNN on objects scaled up 50\%.  (the probability of judging
    $n$ as $m$).}
    \label{table:random_large}
    \begin{footnotesize}
      \begin{tabular}{c|cccccc}
        \toprule
        \diagbox[width=1cm]{$n$}{$m$} & 1 & 2 & 3 & 4 & 5 & 6 \\
        \midrule
        1 & 0.991 & 0.009 & 0 & 0 & 0 & 0 \\
        2 & 0.016 & 0.984 & 0 & 0 & 0 & 0 \\
        3 & 0 & 0.504 & 0.496 & 0 & 0 & 0 \\
        4 & 0 & 0 & 0.793 & 0.207 & 0 & 0 \\
        5 & 0 & 0 & 0.002 & 0.966 & 0.032 & 0 \\
        6 & 0 & 0 & 0 & 0.114 & 0.860 & 0.026 \\
        \bottomrule
      \end{tabular}
    \end{footnotesize}
\end{table}
  % \hfill
\begin{table}[!ht]
    \centering
    \caption{Performance of the subitizing DCNN on objects scaled down 50\% (the probability of judging
    $n$ as $m$).}
    \label{table:random_small}
    \begin{footnotesize}
      \begin{tabular}{c|cccccc}
        \toprule
        \diagbox[width=1cm]{$n$}{$m$} & 1 & 2 & 3 & 4 & 5 & 6 \\
        \midrule
        1 & 0.687 & 0.313 & 0 & 0 & 0 & 0 \\
        2 & 0.026 & 0.390 & 0.583 & 0.001 & 0 & 0 \\
        3 & 0.002 & 0.006 & 0.021 & 0.896 & 0.075 & 0 \\
        4 & 0 & 0 & 0 & 0.014 & 0.492 & 0.494 \\
        5 & 0 & 0 & 0 & 0.001 & 0.043 & 0.956 \\
        6 & 0 & 0 & 0 & 0 & 0.012 & 0.988 \\
        \bottomrule
      \end{tabular}
    \end{footnotesize}
\end{table}

However, the improvement of generalization made by all the efforts above still does not suffice to match the human performance on subitizing.   The accuracy of the newly trained subitizing DCNN deteriorates significantly if we merely scale the objects in test images.  Tables~\ref{table:random_large} and \ref{table:random_small} show what happens if the objects in test images are scaled by $50\%$ up or down, respectively.  The DCNN underestimates (overestimates) $n$ when objects in a test image are scaled up (down).  This seems quite counter-intuitive at first glance, but then the failure to generalize in object size can be explained; the DCNN apparently correlates the class label $n$ to the likely number of boundary pixels per object.

Recall that we have made the class label $n$ statistically independent of the total number of edge points $c$ (see Fig.~\ref{fig:hist_area}) in the training images; otherwise, the DCNN will escape to exploit the positive correlation between $n$ and $c$, instead of understanding the true topological nature of the problem. Here we face a dilemma: normalizing $c$ against $n$ creates a negative correlation between $n$ and $\bar{c}$, the average number of edge points per object; normalizing $\bar{c}$ against $n$ creates a positive correlation between $n$ and $c$.  Because it is impossible to make both $c$ and $\bar{c}$ independent of $n$, the inevitable conclusion is that no preparation of the training data can prevent the DCNN from making a statistical inference of $n$ and force it to discover the essence of connected components underlying subitizing.

\textbf{Other networks.}  We also try to improve the DCNN of Table~\ref{table:dcnn} by training networks with residual blocks for subitizing.
However, the residual DCNN cannot generalize any better than the network of Table~\ref{table:dcnn}.

\section{5. Deterministic General DCNN Solution}

The previous section exposes the handicap of DL on subitizing, a rudimentary cognitive function for humans.  But the exposed deficit of the black box connectionist AI approach on subitizing does not mean that DCNNs cannot do the job.  On the contrary, we can construct a deterministic DCNN algorithm of subitizing that can abstract from any shape and size of objects.

The proposed DCNN algorithm is one of mathematical morphology.
It first reduces (abstracts) objects of complex shapes to single pixels by recursively removing boundary pixels of the objects in parallel.  Then the cognition of subitizing trivially follows.  Here all objects are assumed to have no holes in them.

\begin{figure}[!ht]
    \centering
    \includegraphics[width=0.4\textwidth]{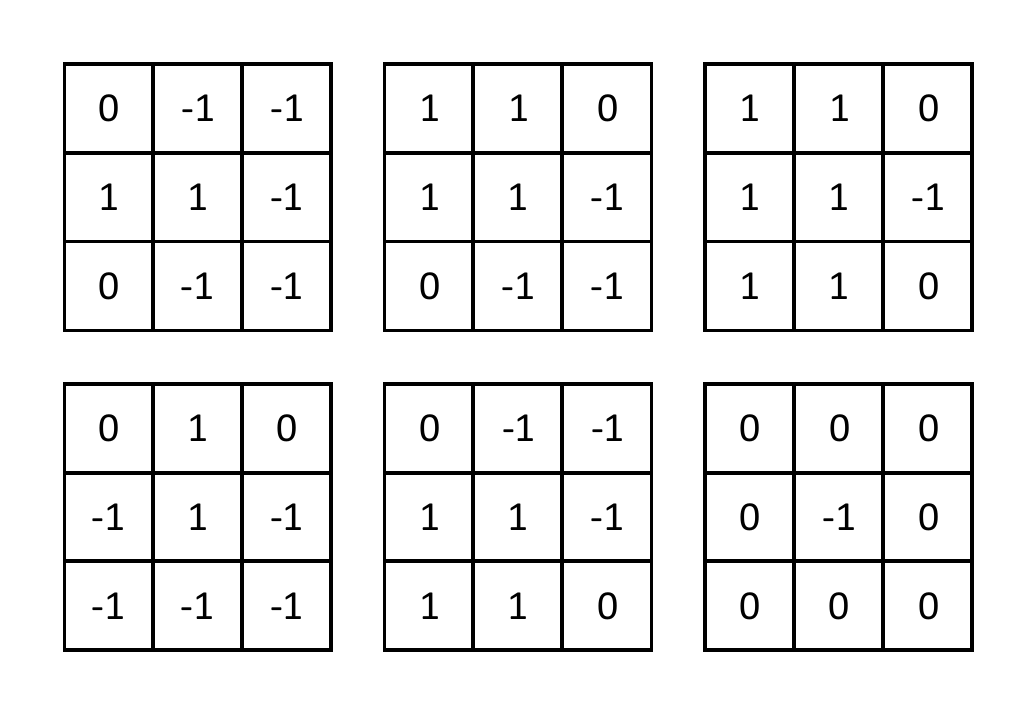}
    \caption{Kernels of the proposed deterministic RCNN algorithm, where $1$,
      $-1$ and $0$ match foreground, background and
      arbitrary pixels, respectively}
    \label{fig:kernels}
\end{figure}
  % \hfill
\begin{figure}[!t]
    \centering
    \includegraphics[width=0.4\textwidth]{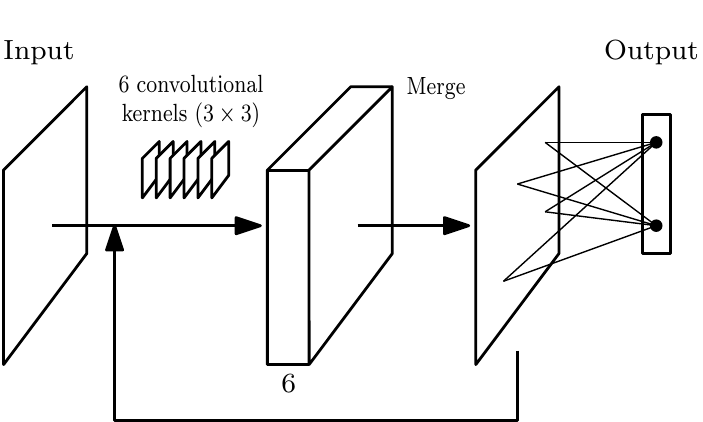}
    \caption{The architecture of the proposed RCNN algorithm.}
    \label{fig:rcnn}
\end{figure}

At the core of the artificial neural network is a small set of $3 \times 3$ convolutional kernels (templates), as listed in Fig.~\ref{fig:kernels}.  If a pixel and its neighbourhood match any of the convolutional kernels, then it can be safely pruned along with all other so matched pixels, without altering the connectivity of any object in the image.  By applying the set of kernels and their 180$^\circ$-rotated versions alternately, the algorithm repeats the pruning process till no further reduction is possible.  This idea can be compactly carried out by a recurrent convolutional neural network (RCNN) with a single conv layer of $6 \times 3 \times 3 = 54$ weights, as illustrated in Fig.~\ref{fig:rcnn}.  Note that in the looping of RCNN, although the geometry of the image signal keeps simplifying, the topology of the objects is preserved, thus maintaining the object count an invariant.  The above connectivity-preserving reduction is the key to achieve the generalization of our solution in object size, which black box DL cannot as reported in Tables~\ref{table:random_large} and \ref{table:random_small}.

What should be stressed is that the proposed RCNN draws the abstraction from an image to object count in a way similar to visual cortex \cite{Hubel1968}, basing its visual cognition on boundary coherence and spatial togetherness \cite{Spelke}.  Moreover, the architecture of our RCNN has a biological parallel.  Indeed, it is well known that the ventral visual pathway contains both feedforward and feedback connections \cite{Felleman1991,Sporns2004,Markov2014}.  Functionally and anatomically feedback connections are vital to visual cognition.

Algorithmically, the proposed RCNN has only 54 weights, whereas other DCNNs, such as the AlexNet and VGGNet, have millions or more.  As expected in the principle of Occam's razor, our RCNN model can abstract from visual signal variations far better than other DCNNs, unfettered by the limitation of i.i.d.~statistical inference as in mainstream machine learning.  Still the proposed RCNN subitizing algorithm has one limitation: it will not work if any object has hole(s) in it.

\section{6. Learning with Preexisted Cognitive Implements}

In the study of human cognitive development, a prevailing hypothesis is that learning or knowledge acquisition is anchored on a small set of core inborn knowledge systems pertaining to domain-specific representational priors; they direct and modulate the learning of novel representations \cite{Spelke,Carey}.  In this view cultural learning is facilitated by a ``cortical recycling'' of a limited number of cerebral circuits biologically evolved to function in ways critical for the survival of our specie.  Furthermore, these elementary cerebral circuits enjoy a sufficient level of plasticity so that their coding scheme can adapt for learning new functions
\cite{Dehaene2007}.

In a similar perspective of nurturing the nature, we can facilitate DL by some preexisted cognitive implements (nature), so it can acquire the capability of performing a specific function, using pertaining training data (nurture).  In our case study of subitizing, we assume that the RCNN of Fig.~\ref{fig:rcnn} together with the accompanying $3 \times 3$ convolutional kernel structure is pre-implemented neural circuitry suited for the task.  Within the given RCNN architecture, the objective of DL reduces to determine the weights of the six convolutional kernels using training images.

We carried out the training of the RCNN with different random initializations of the weights and examine if the weights of the six $3\times 3$ convolutional kernels can be learnt by using the gradient descent method.
The experimental results turn out to be disappointing; the training losses oscillate without exhibiting a downward trend.  To promote the convergence, we strengthen the prior knowledge and initialize the subitizing RCNN not randomly but with the weights only slightly deviated from those of the known solution in Fig.~\ref{fig:kernels}.  Even so, the training still fails to converge; in fact, through the iterations the weights diverge from those of our designed kernels that can solve the problem exactly.  The reason for the failures is that the loss function, given the deterministic RCNN architecture, is highly discontinuous near the solution point, very much like in integer programming.  Although the predetermined RCNN is a suitable connectionist machinery for the cognitive task, its parameters cannot be correctly set by the back propagation algorithm of DL.  This exposes, in our view, a serious handicap of DL due to its optimization methodology of the variational calculus. Many cognitive problems are singular right at the solution points as in the above example.

\begin{figure}[!t]
\centering
\includegraphics[width=0.4\textwidth]{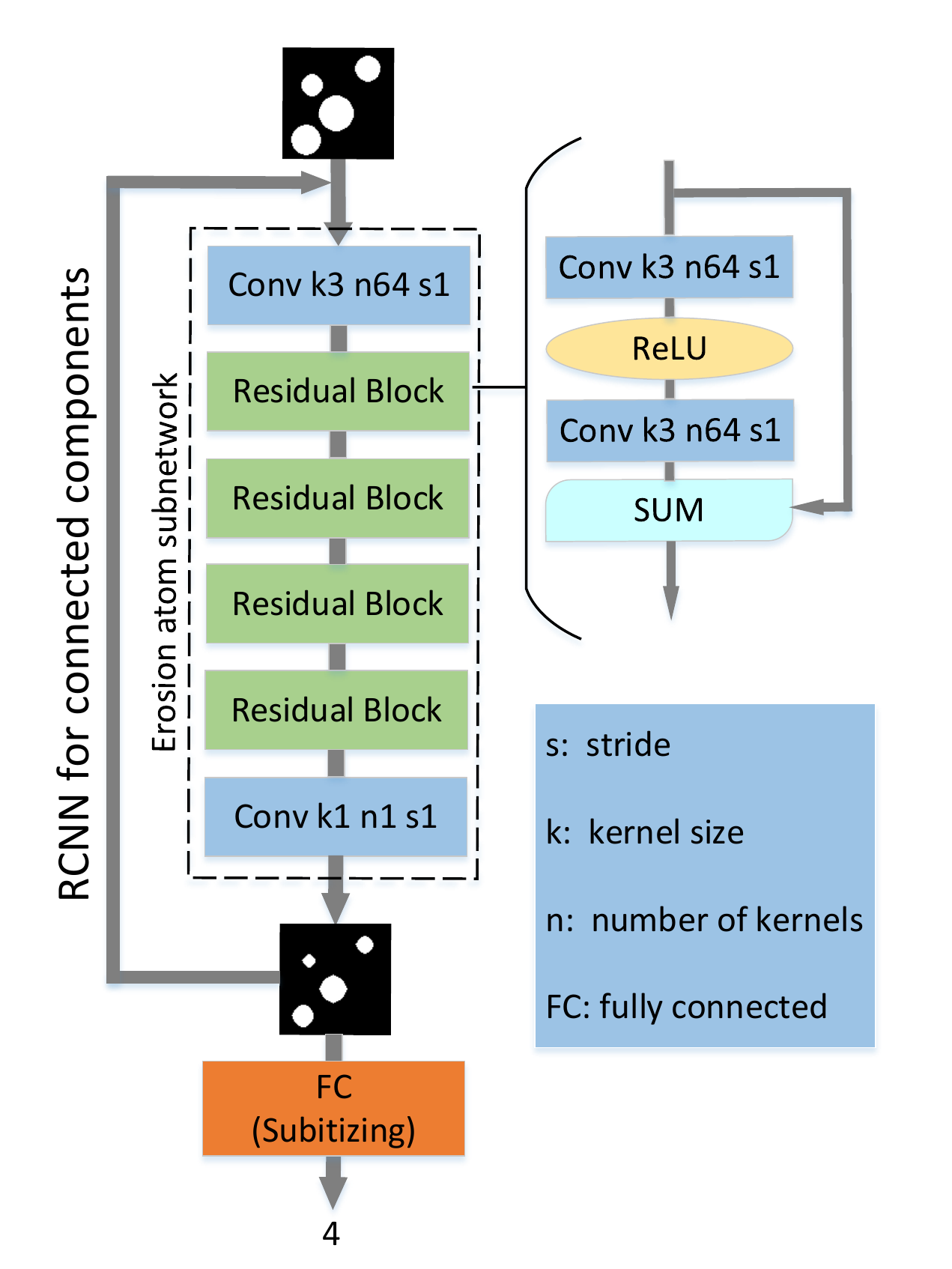}
\caption{The RCNN architecture of connected components and the inner atom subnetwork for erosion.}
\label{fig:erosion}
\end{figure}

Next, we reduce the cognitive task from subitizing to identifying connected components; the latter is the essential feature of the former. If the DL approach can solve the problem of connected components, then the resulting DCNN of connected components can be concatenated to a fully connected network to solve the problem of subitizing as shown in Fig.~\ref{fig:erosion}.

The computation of connected components is a recursive invocation of the morphological atom operation of erosion, or connectivity-preserving reduction. Recalling that the architecture of RCNN has a biological parallel in the ventral visual pathway \cite{Felleman1991,Sporns2004,Markov2014}, we choose the RCNN as the connectionist model for connected components.
In this RCNN the atom operation is the erosion by removing one layer of boundary pixels; this is performed by a fully convolutional neural network as depicted in Fig.~\ref{fig:erosion}.  In our design, the erosion atom subnetwork contains 4 residual blocks. Each residual block has two convolutional layers and one RELU activation layer.

By manually decomposing the cognitive function of subitizing into three subfunctions, one layer erosion, connected component, and counting, we are able to construct subnetworks, one for each subfunction, via separate learning processes, and then combine them into a subitizing DCNN that can generalize to different sizes and shapes of the objects.

Even we have finally succeeded in developing a DCCN solution for subitizing, our success is not very satisfying because it is hardly a case for the power of DL in autonomous cognition. DL accomplishes the cognitive task of subitizing only after we provide the DCNN some predetermined representations and primitives. In other words, it benefits from human insight, far away from the pure end-to-end data driven machine learning.

The above exercise in cognitive computing strongly suggests the necessity of augmenting the black box DL methodology by preexisted (innate) cognitive constructs possessed by humans.

\section{7. Conclusions}

The black box data-driven methodology of DL, specifically supervised learning with DCNNs, is scrutinized on the simple cognitive task of numerosity.  In our carefully controlled experiments DCNNs failed to achieve number abstraction from a large set of training images that exhibit the concept of natural numbers unequivocally. Even with strong supervision, DCNNs did poorly on the tests of subitizing, which children can pass with speed, accuracy and assurance. In contrast, we are able to construct a simple and compact recurrent convolutional neural network that can deterministically perform subitizing.  This work adds a fresh anecdote to the widely held cautions and critiques about the absence of human-like cognitive power of DL.  But on the other hand, if DL is allowed to build upon preexisted cognitive implements, it can learn the abstract notion of numbers.

\bibliography{AI_numerosity_bib}
\bibliographystyle{aaai}

\end{document}